\documentclass[10pt,twocolumn,letterpaper]{article}

\usepackage{cvpr}
\usepackage{times}
\usepackage{epsfig}
\usepackage{graphicx}
\usepackage{amsmath}
\usepackage{amssymb}
\usepackage{caption} % add for teaser image
\usepackage{color}

\usepackage[pagebackref=true,breaklinks=true,letterpaper=true,colorlinks,bookmarks=false]{hyperref}

\definecolor{myPurple}{rgb}{0.4, .0, .8}
\definecolor{myGreen}{rgb}{0, .8, .3}
\definecolor{myRed}{rgb}{0.8, .2, .2}
\definecolor{myOrange}{rgb}{0.8, 0.45, 0.0}
\definecolor{myBlue}{rgb}{.0, .0, 1.0}

\cvprfinalcopy % *** Uncomment this line for the final submission

\def\cvprPaperID{4593} % *** Enter the CVPR Paper ID here

\ifcvprfinal\fi

\title{FaceScape: a Large-scale High Quality 3D Face Dataset and \\ Detailed Riggable 3D Face Prediction}

\author{
Haotian Yang$^{*1}$ \quad Hao Zhu$^{*1,2,5}$ \quad Yanru Wang$^{1}$ \quad Mingkai Huang$^{1}$ \\
\quad Qiu Shen$^{1}$ \quad Ruigang Yang$^{2,3,4,5}$ \quad Xun Cao$^{1}$\\
$^{1}$Nanjing University \quad 
$^{2}$Baidu Research \quad
$^{3}$University of Kentucky \quad
$^{4}$Inceptio Inc. \\
$^{5}$National Engineering Laboratory for Deep Learning Technology and Applications, China\\
}

\begin{document}
\twocolumn[{%
\renewcommand\twocolumn[1][]{#1}%

\maketitle
\vspace{-0.25in}
\begin{center}
    \includegraphics[width=1.0\linewidth]{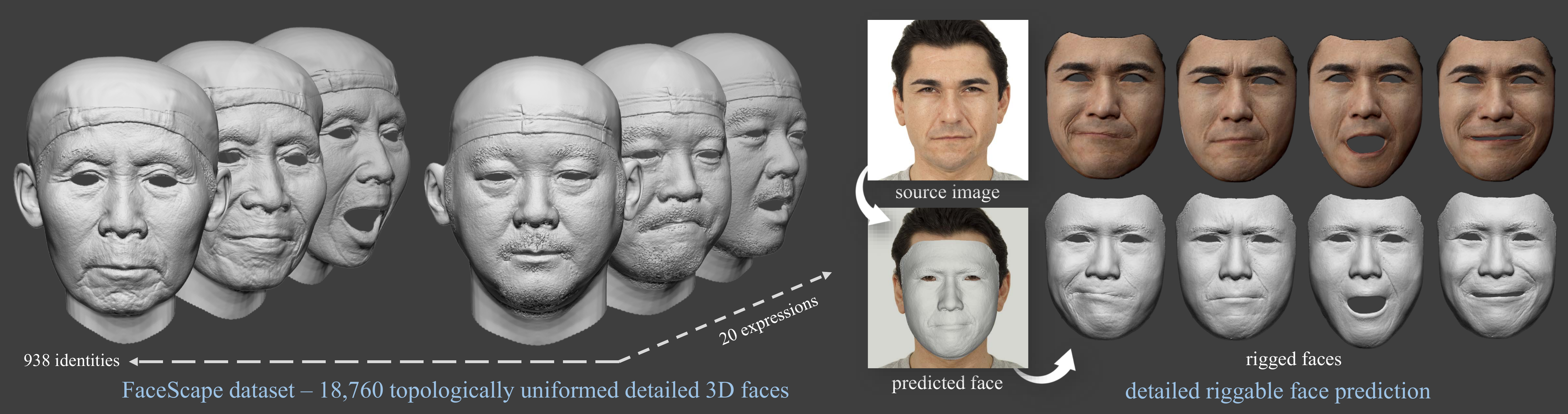}
    \vspace{-0.2in}
    \captionof{figure}{We present FaceScape, a large-scale detailed 3D face dataset consisting of 18,760 textured 3D face models with pore-level geometry.  
    %The faces are captured from 930 subjects and each subject in 20 specified expressions. 
    By learning dynamic details from FaceScape, we present a novel algorithm to predict from a single image a detailed \textbf{rigged} 3D face model that can generate various expressions with high geometric details.}
\label{fig:teaser}
\end{center}
}]

\let\thefootnote\relax\footnotetext{\vspace{-1pt}$*$ These authors contributed equally to this work.}
\let\thefootnote\relax\footnotetext{$\dag$ \url{https://github.com/zhuhao-nju/facescape.git}}

%%%%%%%%% ABSTRACT
\begin{abstract}
In this paper, we present a large-scale detailed 3D face dataset, \emph{FaceScape}, and propose a novel algorithm that is able to predict elaborate riggable 3D face models from a single image input. 
% dataset
FaceScape dataset provides 18,760 textured 3D faces, captured from 938 subjects and each with 20 specific expressions. The 3D models contain the pore-level facial geometry that is also processed to be topologically uniformed. These fine 3D facial models can be represented as a 3D morphable model for rough shapes and displacement maps for detailed geometry.
% method
Taking advantage of the large-scale and high-accuracy dataset, a novel algorithm is further proposed to learn the expression-specific dynamic details using a deep neural network. The learned relationship serves as the foundation of our 3D face prediction system from a single image input. Different than the previous methods, our predicted 3D models are riggable with highly detailed geometry under different expressions. 
%Prior works have achieved to predict detailed shape in a static expression, while the result shapes cannot be rigged naturally as the details are varying across expressions.  In this paper, we proposed to learn the dynamic details from FaceScape dataset using a deep neural network.  A full pipeline is proposed to predict a realistically 3D face which can be rigged to various expressions with plausible detailed shape. 
%which can be applied in animation generation and photo-real image manipulation.  
% public 
The unprecedented dataset and code will be released to public for research purpose$^{\dag}$.
\end{abstract}

%%%%%%%%% BODY TEXT
\section{Introduction}

%Face dataset is usefull
Parsing and recovering 3D face models from images have been a hot research topic in both computer vision and computer graphics due to its many applications. As learning based methods have become the mainstream in face tracking, recognition, reconstruction and synthesis, 3D face datasets becomes increasingly important.  While there are numerous 2D face datasets, the few 3D datasets lack in 3D details and scale. As such, learning-based methods that rely on the 3D information suffer.  

%Existing Face dataset
Existing 3D face datasets capture the face geometry using sparse camera array\cite{booth20163d, dai20173d, phillips2005overview} or active depth sensor such as Kinect\cite{cao2013facewarehouse} and coded light\cite{paysan20093d}.  These setups limit the quality of the recovered faces.  We captured the 3D face model using a dense 68-camera array under controlled illumination, which recovers the 3D face model with wrinkle and pore level detailed shapes, as shown in Figure~\ref{fig:teaser}.  In addition to shape quality, our dataset provides considerable amount of scans for study.  We invited 938 people between the ages of 16 and 70 as subjects, and each subject is guided to perform 20 specified expressions, generating 18,760 high quality 3D face models.  The corresponding color images and subjects' basic information (such as age and gender) are also recorded.

% Bilinear model with displacement map
Based on the high fidelity raw data, we build a powerful parametric model to represent the detailed face shape.  All the raw scans are firstly transformed to a topologically uniformed base model representing the rough shape and a displacement map representing detailed shape. The transformed models are further used to build bilinear models in identity and expression dimension.  Experiments show that our generated bilinear model exceeds previous methods in representative ability.

%The transformed models efficiently represent the detailed shapes as a 3D morphable model with minimal geometry loss, and are ready to build riggable model like blendshapes.  

% riggable model
Using FaceScape dataset, we study how to predict a detailed riggable  face model from a single image.  Prior methods are able to estimate rough blendshapes where no wrinkle and subtle features are recovered.  The main problem is how to predict the variation of small-scale geometry caused by expression changing, such as wrinkles.  We propose the dynamic details which can be predicted from a single image by training a deep neural network on FaceScape dataset. Cooperated with bilinear model fitting method, a full system to predict detailed riggable model is presented.  Our system consists of three stages: base model fitting, displacement map prediction and dynamic details synthesis.  
%The key novelty is that geometry details are disentangled into static details and dynamic details using a deep neural network. 
%Given a single input image, a base model is fitted using classic morphable model techniques. In displacement prediction phase, the dynamic details are learned from the FaceScape dataset using a deep neural network, which help synthesize realistic details for rigged model.  
As shown in Figure~\ref{fig:teaser}, our method predicts detailed 3D face model which contains subtle geometry, and achieves high accuracy due to the powerful bilinear model generated from FaceScape dataset.  The predicted model can be rigged to various expressions with plausible detailed geometry.  
%The predicted model can further be used in photo-real facial image manipulation.
%A benchmark is also provided to evaluate 3D face reconstruction accuracy.

Our contributions are summarized as following:
\begin{itemize} 
	\item We present a large-scale 3D face dataset, FaceScape, consisting of 18,760 extremely detailed 3D face models.  All the models are processed to topologically uniformed base models for rough shape and displacement maps for detailed shape.  The data are released free for non-commercial research.

	%\item The displacement mapping and 3D morphable model are combined to efficiently represent the detailed facial geometry while keeping model morphable.  
	%\red{(ryang: can you say some numbers, such as the new representation takes X\% of the original data size, while maintaining the error to be XX mm with the raw data? )}
	
	\item We model the variation of detailed geometry acrossing expressions as dynamic details, and propose to learn the dynamic detail from FaceScape using a deep neural network.
	
	\item A full pipeline is presented to predict detailed riggable face model from a single image.  Our result model can be rigged to various expressions with plausible geometric details.

\end{itemize} 

\section{Related Work}
\label{sec:related}
\noindent \textbf{3D Face Dataset.}
3D face datasets are of great value in face-related research areas.  Existing 3D face datasets could be categorized according to the acquisition of 3D face model.  Model fitting datasets\cite{paysan20093d, zhu2016face, guo2018cnn, booth20173d, booth20183d} fit the 3D morphable model to the collected images, which makes it convenient to build a large-scale dataset on the base of wild faces.  The major problem of the fitted 3D model is the uncertainty of accuracy and the lack of detailed shape.  To obtain the accurate 3D face shape, a number of works reconstructed the 3D face using active method including depth sensor or scanner\cite{yin20063d, yin126high, baocai2009bjut, savran2008bosphorus, sankowski2015multimodal, cao2013facewarehouse, cheng20184dfab}, while the other works built sparse multi-view camera system\cite{zhang2014bp4d, Cosker:2011:FVD:2355573.2356405}.  Traditional depth sensors and 3D scanners suffer from the limited spatial resolution, so they can't recover detailed facial geometry.  
%Laser scanners usually require seconds of time for facial scanning, which introduces jitter error as the human are not absolutely static.  
The sparse multi-view camera system suffers from the unstable and inaccurate reconstruction\cite{seitz2006comparison, zhu2017role, zhu2016video}.  The drawbacks of these methods limit the quality of 3D face model in previous datasets.  Different from the datasets above, FaceScape obtained the 3D face model from a dense multi-view system with 68 DSLR cameras, which provides extremely high quality face models.  The parameters measuring 3D model quality are listed in Table~\ref{tab:datasets}.  Our dataset outperforms previous works on both model quality and data amount.  Note that Table~\ref{tab:datasets} doesn't list the datasets which provide only parametric model but no source 3D models\cite{booth20163d, booth2018large, paysan20093d, luthi2017gaussian}.

\begin{table*}[]
\caption{Comparison of 3D Face Datasets}

\begin{tabular}{cccccc}
\hline

Dataset                              & \multicolumn{1}{l}{Sub. Num}  & \multicolumn{1}{l}{Exp. Num} & \multicolumn{1}{l}{Vert. Num} & \multicolumn{1}{l}{Image/Texture Resolution}         & \multicolumn{1}{c}{ Source}          \\ \hline
BU-3DFE\cite{yin20063d}    & 100    & 25    & 10k-20k    & $1300\times900$ / -    & structure light    \\
BU-4DFE\cite{yin126high}   & 101    & 6(video)    & 10k-20k    & $1040\times1329$ / -    & structure light  \\
BJUT-3D\cite{baocai2009bjut}    & 500    & 1-3    & $\approx$200k    & $478\times489$ / -   & laser scanner   \\
Bosphorus\cite{savran2008bosphorus}    & 105    & 35    & $\approx$35k  & $1600\times1200$ / -    & structure light \\
%MorphFace\cite{paysan20093d, luthi2017gaussian}    & 100    & 6   & Yes    & $\approx$100k    &  vertex color  & scanner    \\
%DMCSv1\cite{sankowski2015multimodal}    & 35    & 30  &  --  & $424\times320$ / -    & structure light    \\
%FLAME\cite{li2017learning}    & 200    & \textbf{video}     & No    & $\approx$20k    & $1280\times720$   & scanner    \\
FaceWarehouse\cite{cao2013facewarehouse}    & 150    & 20    & $\approx$11k    & $640\times480$ / -   & kinect    \\
4DFAB\cite{cheng20184dfab}    & 180    & 6(video)    & $\approx$100k    & $1200\times1600$ / -   & kinect+cameras(7)    \\
%UMB-DB\cite{colombo2011umb}    & 143    & 4    &  --  &  --   & scanner    \\
D3DFACS\cite{Cosker:2011:FVD:2355573.2356405}    & 10    & 38AU(video)    & $\approx$30k    &  - / $1024\times1280$   & multi-view system(6)    \\
BP4D-Spontanous\cite{zhang2014bp4d}    & 41    & 27AU(video)    & $\approx$37k    &  $1040\times1392$ / -   &  multi-view system(3)    \\
{\textbf{FaceScape (Ours)}} & \textbf{938}    & 20    & \textbf{$\approx$2m}     & \textbf{4k-8k} / \textbf{4096$\times$4096}    & multi-view system(68)    \\ \hline
\label{tab:datasets}
\end{tabular}
\end{table*}

%JNU-3D\cite{koppen2018gaussian}
%FRGC\cite{phillips2005overview} Vivid 900/910 sensor, 50000 data, in the wild
%FaceWarehouse \cite{cao2013facewarehouse} captures 3D face shape using KinectFusion\cite{newcombe2011kinectfusion}, 
%MeIn3D\cite{booth20163d, booth2018large}  A 3d morphable model learnt from 10,000 faces
%Basel Face Model (BFM)\cite{paysan20093d}
%\cite{feng2018evaluation}
%\cite{dai20173d} 1.5k subjects, A 3dMD five-camera system, 180K vertices joined into typically 360K triangles
%FaceBase\cite{sharif2012enhanced} - Enhanced and fast face recognition by hashing algorithm, no 3D shape
%Size China project \cite{ball20103d} - A 3d anthropometric survey of the chinese head, cannot find the download-able paper

\noindent \textbf{3D Morphable Model.} 3DMM is a statistical model which transforms the shape and texture of the faces into a vector space representation\cite{blanz1999morphable}.  As 3DMM inherently contains the explicit correspondences from model to model, it is widely used in model fitting, face synthesis, image manipulations, etc.  The recent research on 3DMM can be generally divided into two directions.  The first direction is to separate the parametric space to multiple dimensions like identity, expression and visemes, so that the model could be controlled by these attributes separately\cite{vlasic2005face, cao2013facewarehouse, li2017learning, jiang2019disentangled}.  The models in expression dimension could be further transformed to a set of blendshapes\cite{li2010example}, which can be rigged to generate individual-specific animation.  Another direction is to enhance the representation power of 3DMM by using deep neural network to present 3DMM bases \cite{bagautdinov2018modeling, tewari2018self, tran2019towards, tran2019learning, tran2018nonlinear, cheng2019meshgan}.  

%Neural network based 3DMM can be trained from images by weak supervised learning, which 

%One drawback of this non-linear 3DMM is the difficulties to rigging the model in one certain attributes as the 
%Basel\cite{paysan20093d}
%Flame\cite{li2017learning}
%Facewarehouse\cite{cao2013facewarehouse}

\noindent \textbf{Single-view shape Prediction.}
Predicting 3D shape from a single image is a key problem to many applications like view synthesis\cite{flynn2016deepstereo, zhu2018view, zhu2019detailed} and stereoscopic video generation\cite{cao2011semi, huang2014toward}.  The emergence of 3DMM has simplified the single-view reconstruction of face to a model fitting problem, which could be well solved by fitting facial landmarks and other features\cite{romdhani2005estimating, thies2016face2face} or regressing the parameter of 3DMM with a deep neural network\cite{dou2017end, zhu2016face}.  However, fitting 3DMM is difficult in recovering small details from the input image due to the limited representation power.  To solve this problem, several recent works adopt the multi-layer refinement structures.  
Richardson~\etal \cite{richardson2017learning} and Sela~\etal \cite{sela2017unrestricted} both proposed to firstly predict a rough facial shape and render it to the depth map, then refine the depth map to enhance the details from the registered source image.
Sengupta~\etal \cite{sengupta2018sfsnet} proposed to train the SfSNet on combination of labeled synthetic data and unlabeled in-the-wild data to estimate plausible detailed shape in unconstrained images.
Tran~\etal \cite{tran2018extreme} proposed to predict a bump map to represent the wrinkle-level geometry base on a rough base model.  Huynh~\etal \cite{huynh2018mesoscopic} utilized image-to-image network and super-resolution network to recover the mesoscopic facial geometry in the form of displacement map.  Chen~\etal \cite{chen2019photo} also tried to predict the displacement map with a conditional GAN based on the 3DMM model, which enables to recover detailed shape from an in-the-wild image.

Our work advances the state of the art in multiple aspects. In dataset, our FaceScape is by far the largest with the highest quality. A detailed quantitative comparison with previous datasets are made in Table~\ref{tab:datasets}.  
In 3D face prediction, previous works focus on enhancing the static detailed facial shape, while we study the problem of recovering an \textbf{animable} model from a single image.  We demonstrate for the first time that a detailed and rigged 3D face model can be recovered from a single image. The rigged model exhibits expression-depended geometric details such as wrinkles.  
\section{Dataset}

\subsection{3D Face Capture}

\begin{figure}[t]
\begin{center}
    \includegraphics[width=1.0\linewidth]{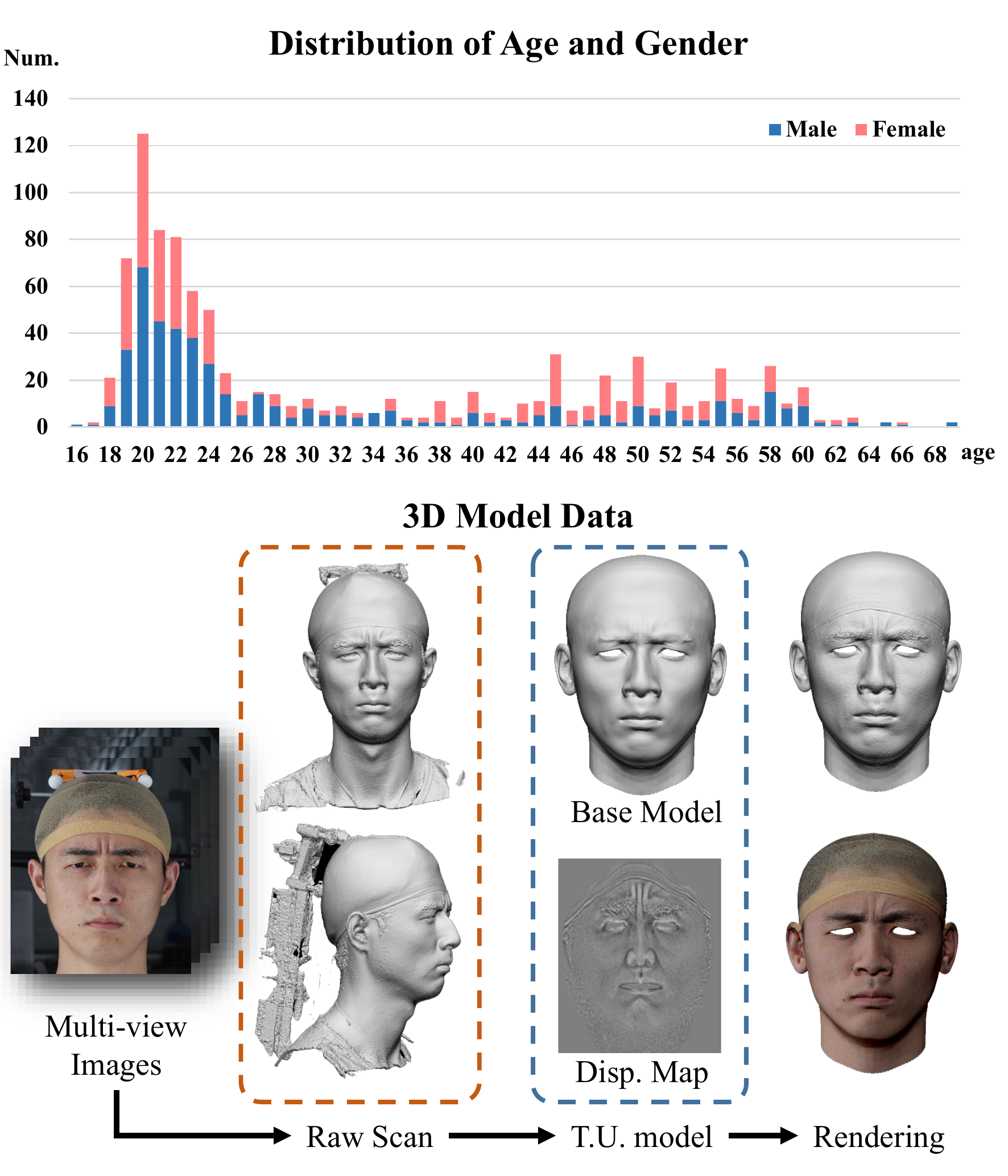}
\end{center}
    \vspace{-0.25in}
    \caption{Description of FaceScape dataset.  In the upper side we show the histogram of subjects' age and gender.  In the lower side we show the pipeline from the captured multi-view images to topologically uniformed models (T.U. models). }
\label{fig:statistic}
\end{figure}

We use a multi-view 3D reconstruction system to capture the raw mesh model for the datasets.  The multi-view system consists of 68 DSLR cameras, 30 of which capture 8K images focusing on front side, and the other cameras capture 4K level images for the side part.  The camera shutters are synced to be triggered within $5$ms.  We spend six months to invite 938 people to be our capturing subjects.  The subjects are between 16 and 70 years old, and are mostly from Asia.  We follow FaceWarehouse\cite{cao2013facewarehouse} which asks each subject to perform 20 specific expressions including neutral expression for capturing.  The total reconstructed number reach to roughly 18,760, which is the largest amount comparing to previous expression controlled 3D face datasets.  The reconstructed model is triangle mesh with roughly 2 million vertices and 4 million triangle faces.
%As the expressions for each subjects are specified, these models could be used to build blendshapes for future usage.  
The meta information for each subject is recorded, including age, gender, and job (by voluntary).  We show the statistical information about the subjects in our dataset in Figure~\ref{fig:statistic}, and a comparison with prior 3D face datasets in Table~\ref{tab:datasets}.

 %Though the huge amount of vertices express the detailed geometry, they are too heavy to manipulate, and the geometry correspondence across models are uncertain.  
%\red{A widely used solution is to build a 3D morphable model\cite{blanz1999morphable}, which represents all 3D faces in a topology-uniformed mesh with registered geometry features. }

\subsection{Topologically Uniformed Model}

% 3DMM is .........  With the 3DMM .......
% Different from previous dataset, building 3DMM for our dataset require a large vertex/face space to represent detailed geometry.
% We treat the face model in two stage

%\textbf{3DMM for base shape.} 
We down-sample the raw recovered mesh into rough mesh with less triangle faces, namely base shape, and then build 3DMM for these simplified meshes.  Firstly, we roughly register all the meshes to the template face model by aligning 3D facial landmarks, then the NICP\cite{amberg2007optimal} is used to deform the templates to fit the scanned meshes.  The deformed meshes can be used to represent the original scanned face with minor accuracy loss, and more importantly, all of the deformed models share the uniform topology. The detailed steps to register all the raw meshes are described in the supplementary material.

%We firstly extract 2D landmarks from the frontal image, then get the corresponding 3D landmarks by inverse-projection 2D landmarks.  The Procrustes transformation\cite{gower1975generalized} is used to register all landmarks to a standard 3D facial template with landmark annotations.  In this way, the pose and scale for all the scanned mesh are roughly aligned to the standard facial template.  Then we directly use Non-rigid ICP\cite{amberg2007optimal} to register the standard template mesh to scanned mesh in neutral expression. For scanned meshes in other 19 expressions, similar to \cite{cao2013facewarehouse}, we first use deformation transfer algorithm\cite{sumner2004deformation} to deform the registered mesh in neutral expression to other expressions mimicking the deformation of a set of template meshes in corresponding expression. Then we use NICP to register these deformed user specific templates to scanned meshes to fit the scans in non-neutral expression more accurately.

%\textbf{Representation of detail.} 
After obtaining the topology-uniformed base shape, we use displacement maps in UV space to represent middle and fine scale details that are not captured by the base model due to the small number of vertices and faces. 
%The most straightforward way to compute the displacement map is to calculate the distance from the surface of the registered model to the raw mesh. However, we find that there will be artifacts in the displacement map caused by the defects in the registration procedure. Thus 
%We firstly smooth the raw scan using Laplacian mesh smoothing, then 
We find the surface points of base mesh corresponding to the pixels in the displacement map, then inverse-project the points to the raw mesh along normal direction to find its corresponding points. The pixel values of the displacement map is set to the signed distance from the point on base mesh to its corresponding point.  
% This is too detailed 
%We also test more sophisticated mesh smoothing methods such as HC Laplacian smooth \cite{vollmer1999improved} and Taubin smooth\cite{taubin1995signal} and find the simple Laplacian smooth is sufficient.

We use base shapes to represent rough geometry and displacement maps to represent detailed geometry, which is a two-layer representation for our extremely detailed face shape.  The new representation takes roughly $2\%$ of the original mesh data size, while maintaining the mean absolute error to be less than $0.3$mm.

%\zhuhao{3.2:topology uniformed model  3.3:3DMM}
\subsection{Bilinear Model}\label{sec:3dmm}
%\zhuhao{the below is newly written, please review. @haotian}
Bilinear model is firstly proposed by Vlasic~\etal~\cite{vlasic2005face}, which is a special form of 3D morphable model to parameterize face models in both identity and expression dimensions. The bilinear model can be linked to a face-fitting algorithm to extract identity, and the fitted individual-specific model can be further transformed to riggable blendshapes. 
Here we describe how to generate bilinear model from our topologically uniformed models.  %3D Morphable Model (3DMM) is the parametric model representing the face shape in a parametric space, which is essential for riggable model prediction method.
%\zhuhao{hao will modify this subsection, uniform the usage of 3DMM}
%In contrast to previous datasets\cite{cao2013facewarehouse, li2017learning}, the raw recovered models in FaceScape are extremely detailed, and such a large amount of vertices is difficult to be expressed in a single mesh based 3DMM.  Therefore, we express the facial geometry in a dual-layer model.  In the first layer, the rough geometry is expressed by a base 3DMM with 26317 vertices.  In the second layer, the displacement map is built to express the residual geometry based on the base model.
%Riggable model is 3DMM which can be rigged to different expressions while keep the identity fixed. This requires to disentangled the parametric space into expression dimension and identity dimension.  
Given 20 registered meshes in different expressions, we use the example based facial rigging algorithm\cite{li2010example} to generate 52 blendshapes based on FACS\cite{Ekman1978FacialAC} for each person. Then we follow the previous methods\cite{vlasic2005face, cao2013facewarehouse} to build the bilinear model from generated blendshapes in the space of 26317 vertices $\times$ 52 expressions $\times$ 938 identities. Specifically, we use Tucker decomposition to decompose the large rank-3 tensor to a small core tensor $C_r$ and two low dimensional components for identity and expression. New face shape can be generated given the the identity parameter $\textbf{w}_{id}$ and expression parameter $\textbf{w}_{exp}$ as:
\begin{equation}
\begin{aligned}
\mathop{V}=C_r \times \textbf{w}_{exp} \times \textbf{w}_{id}
\end{aligned}
\end{equation}
where $V$ is the vertex position of the generated mesh. 
%We perform the decomposition only along the identity dimension to reserve the semantic meaning of parameters in expression dimension as specific blendshape weights in the rest of the paper, unless specified.

\begin{figure}[t]
\begin{center}
    \includegraphics[width=1.0\linewidth]{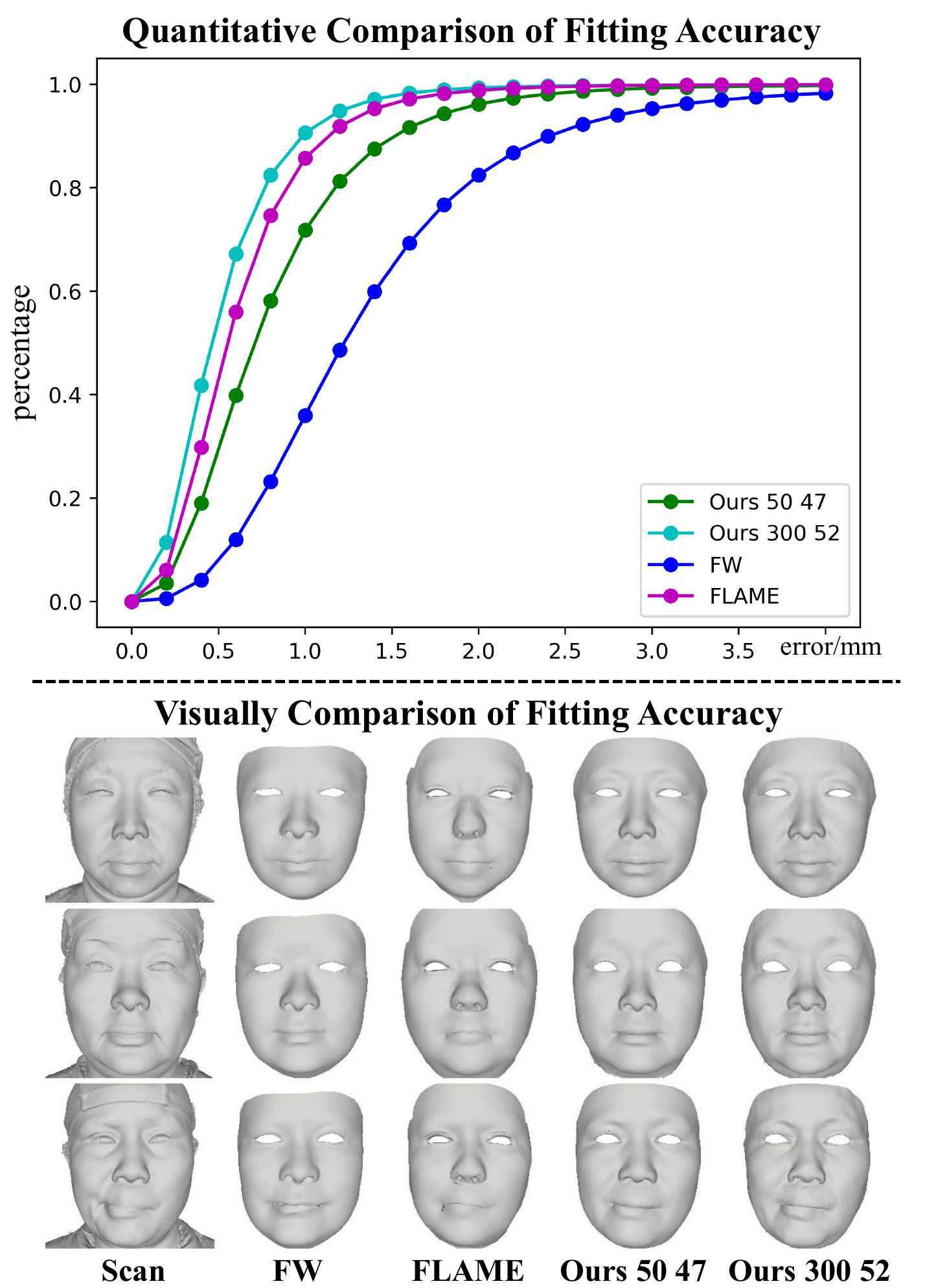}
\end{center}
    \vspace{-0.2in}
    \caption{Comparison of Reconstruction Error for parametric model generated by FaceScape and previous datasets.}
\label{fig:representation}
\end{figure}

The superiority in quality and quantity of FaceScape makes the generated bilinear model own higher representation power.  We evaluate the representation power of our model by fitting it to scanned 3D meshes not part of the training data. We compare our model to FaceWarehouse(FW)\cite{cao2013facewarehouse} and FLAME\cite{li2017learning} by fitting them to our self-captured test set, which consists of 1000 high quality meshes from 50 subjects performing 20 different expressions each. FW has 50 identity parameters and 47 expression parameters, so we use the same number of parameters for fair comparison. To compare with FLAME which has 300 identity parameters and 100 expression parameters, we use 300 identity parameters and all 52 expression paremeters. Figure~\ref{fig:representation} shows the cumulative reconstruction error. Our bilinear face model achieves much lower fitting error than FW using the same number of parameters and also outperform FLAME using even less expression parameters. The visually comparison in Figure~\ref{fig:representation} shows ours model could produce more mid-scale details than FW and FLAME, leading to more realistic fitting results.

%\noindent \textbf{Detailed shape representation.}  
%Similar to the previous works ~\cite{huynh2018mesoscopic} and ~\cite{chen2019photo}, we use a displacement map to represent the detailed geometry on the base of the base shape.

% \noindent \textbf{Texture.} 
% We blend the multi-view image to build the UV texture mapping for the base mesh model.  \textcolor{red}{todo: texture generating method.}  The UV texture image is 4K level, which expressed the very detailed photometric features on the face.  
\section{Detailed Riggable Model Prediction}

\begin{figure}[t]
\begin{center}
    \includegraphics[width=1.0\linewidth]{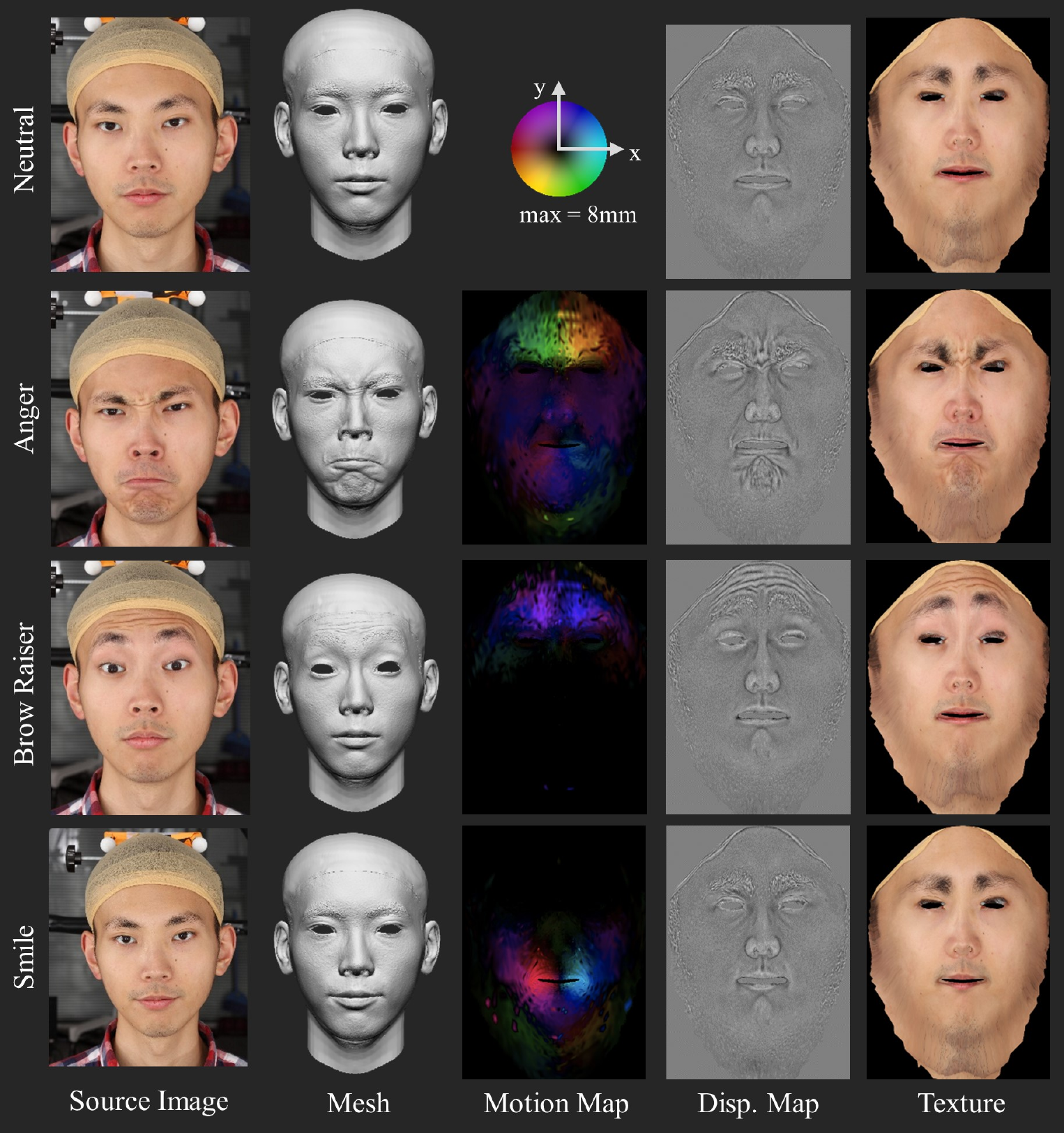}
\end{center}
    \vspace{-0.2in}
    \caption{Riggable details can be decoupled as static details and dynamic details.  The static details can be estimated from the facial textures, while the dynamic details are strongly related to the facial deforming map.
    %\red{(replace the last row with another example)} 
    }
\label{fig:detail}
\end{figure}

As reviewed in the related works in Section~\ref{sec:related}, existing methods have succeed in recovering extremely detailed 3D facial model from a single image.  However, these recovered models are not riggable in expression space, since the recovered detail is static to the specific expression.  Another group of works try to fit a parametric model to the source image, which will obtain an expression-riggable model, but the recovered geometry stays in the rough stage.  

\begin{figure*}[t]
\begin{center}
    \includegraphics[width=1.0\linewidth]{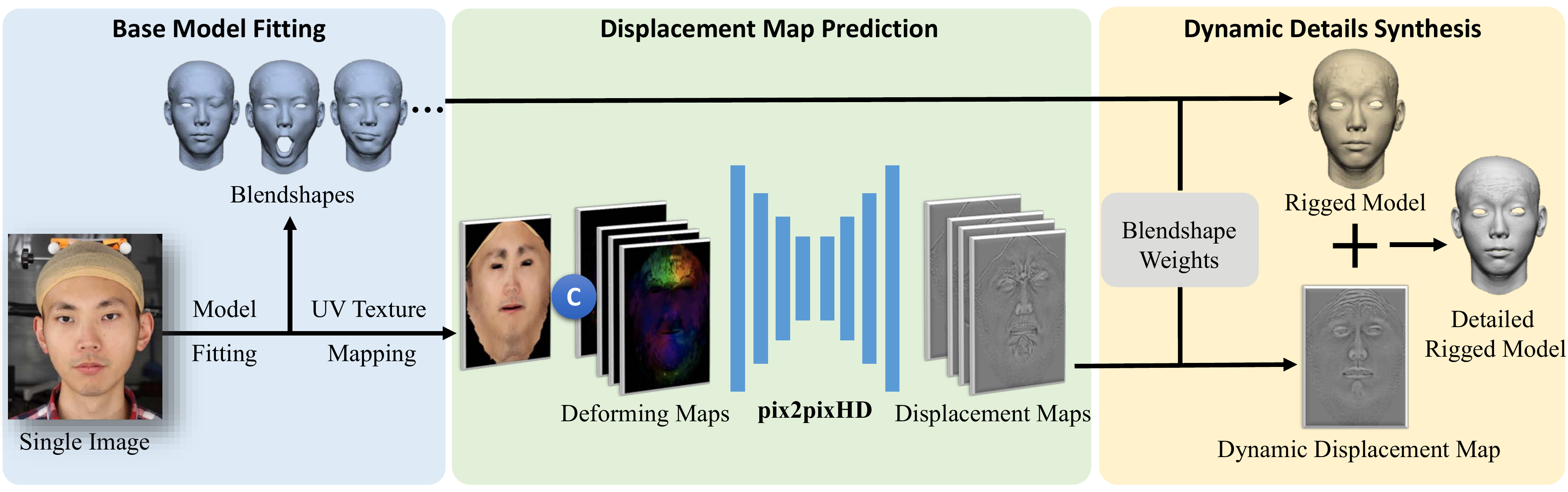}
    \vspace{-0.4in}
\end{center}
    \vspace{-0.02in}
    \caption{The pipeline to predict a detailed riggable 3D face from a single image consists of three stages: base model fitting, displacement map prediction, and dynamic details synthesis. %\zhuhao{last stage to update, name change to `Dynamic Details Synthesis'}
    }
\label{fig:method}
\end{figure*}

The emerge of FaceScape dataset makes it possible to estimate detailed and riggable 3D face model from a single image, as we can learn the dynamic details from the large amount of detailed facial models.  We show our pipeline in Figure~\ref{fig:method} to predict a detailed and riggable 3D face model from a single image.  The pipeline consists of three stages: base model fitting, displacement map prediction and dynamic details synthesis.  We will explain each stage in detail in the following sections.

\subsection{Base Model Fitting}
The bilinear model for base shape is inherently riggable as the parametric space is separated into identity dimension and expression dimension, so the rough riggable model can be generated by regressing the parameters of identity for the bilinear model. Following \cite{thies2016face2face}, we estimate parameters corresponding to a given image by optimizing an objective function consisting of three parts.
The first part is landmark alignment term. Assuming the camera is weak perspective, the landmark alignment term $E_{lan}$ is defined as the distance between the detected 2D landmark and its corresponding vertex projected on the image space. The second part is pixel-level consistency term $E_{pixel}$ measuring how well the input image is explained by a synthesized image. The last part is regularization term which formulates identity, expression, and albedo parameters as multivariate Gaussians.
The final objective function is given by:
\begin{equation}
\begin{aligned}
E=E_{lan}+\lambda_{1} E_{pixel}+{\lambda_2 E_{id}}+{\lambda_3 E_{exp}}+{\lambda_4 E_{alb}}
\end{aligned}
\end{equation}
where $E_{id}$, $E_{exp}$ and $E_{alb}$ are the regularization terms of expression, identity and albedo, respectively. $\lambda_1$, $\lambda_2$, $\lambda_3$ and $\lambda_4$ are the weights of different terms. 

After obtaining the identity parameter $w_{id}$, individual-specific blendshapes $B_i$ can be generated as:
\begin{equation}
\begin{aligned}
B_i=C_r \times \hat{w_{exp}}^{(i)} \times w_{id}, 0\leq i\leq 51
\end{aligned}
\end{equation}
where $\hat{w_{exp}}^{(i)}$ is the expression parameter corresponding to blendshape $B_i$ from Tucker decomposition. 

\subsection{Displacement Map Prediction}
Detailed geometry is expressed by displacement maps for our predicted model.  In contrast to the static detail which is only related to the specific expression in a certain moment, dynamic detail expresses the geometry details in varying expressions.  Since the single displacement map cannot represent the dynamic details, we try to predict multiple displacement maps for 20 basic expressions in FaceScape using a deep neural network.
%Building a bilinear model for the displacement map just like the base model seems like a solution to model the riggabe details, however, the number of pixels on a displacement map is much more than the number of vertices on the base model, which makes it even hard to build a bilinear model with limited computation source.
%Furthermore, the detail variation between identities is difficult to be represented with a simple linear mapping.  
%Therefore, we propose to use a neural network to predict the riggable details, more specifically, to predict displacement maps for 20 basic expressions.

%We firstly explain the difference between the static detail prediction and dynamic detail prediction.  Static details are only related to the expression in the source image, while dynamic details can express the geometry details in all expressions.  This means the dynamic details covers not only current expressions but also other expressions in each blendshape basis.

We observed that the displacement map in a certain expression could be decoupled into static part and dynamic part.  The static part tends to keep static in different expressions, and is mostly related to the intrinsic feature like pores, nevus, and organs.  The dynamic part varies in different expressions, and is related to the surface shrinking and stretching.  We use a deforming map to model the surface motion, which is defined as the difference of vertices' 3D position from source expression to target expression in the UV space.  As shown in Figure~\ref{fig:detail}, we can see the variance between displacement maps is strongly related to the deforming map, and the static features in displacement maps are related to the texture. So we feed motion maps and textures to a CNN to predict the displacement map for multiple expressions.

%The intuition is that the static details are local geometry features that intrinsically exist on the face, like pores, nevus, and shape of organs.  These features can be predicted from the facial textures.  While the dynamic details are  wrinkles generated by the facial motions, which can be predicted from the surface motions.  We used motion map to describe the facial motion, which is the difference of vertices' position between source expression and the to-predict expression in the UV space.

We use pix2pixHD\cite{wang2018high} as the backbone of our neural network to synthesize high resolution displacement maps.  The input of the network is the stack of deforming map and texture in UV space, which can be computed from the recovered base model.  Similar to \cite{wang2018high}, the combination of adversarial loss $L_{adv}$ and feature matching loss $L_{FM}$ is used to train our net with the loss function formulated as:
\begin{equation}
\begin{aligned}
\mathop{\min}_{G} ((&\mathop{\max}_{D_1, D_2, D_3} \sum_{k=1,2,3} L_{adv}(G,D_k)) \\& + \lambda \sum_{k=1,2,3}L_{FM}(G,D_k))
\end{aligned}
\end{equation}
where $G$ is the generator, $D_1, D_2$ and $D_3$ are discriminators that have the same LSGAN\cite{mao2017least} architecture but operate at different scales, $\lambda$ is the weight of feature matching loss.

\subsection{Dynamic Detail Synthesis}
Inspired by\cite{nagano2018pagan}, we synthesize displacement map $F$ for an arbitrary expression corresponding to specific blendshape weight $\boldsymbol{\alpha}$, using a weighted linear combination of generated displacement maps $\hat{F}_0$ in neutral expression and $\hat{F}_i$ in other 19 key expressions:
\begin{equation}
\begin{aligned}
\mathop{F}=M_0 \odot \hat{F}_0 + \sum_{i=1}^{19} M_i \odot \hat{F}_i
\end{aligned}
\end{equation}
where $M$ is the weight mask with the pixel value between $0$ and $1$ , $\odot$ is element-wise multiplication operation. To calculate the weight mask, considering the blendshape expressions change locally, we first compute an activation mask $A_j$ in UV space for each blendshape mesh $e_j$ as:
\begin{equation}
\begin{aligned}
\mathop{A_j(p)}=||e_j(p)-e_0(p)||_2
\end{aligned}
\end{equation}
where $A_j(p)$ is the pixel value at position $p$ of the $j$th activation mask, $e_j(p)$ and $e_0(p)$ is the corresponding vertices position on blendshape mesh $e_j$ and neutral blendshape mesh $e_0$, respectively. The activation masks are further normalized between 0 and 1. Given the activation mask $A_j$ for each of the 51 blendshape meshes, the $i$th weight mask $M_i$ is formulated as a linear combination of the activation masks weighted by the current blendshape weight $\boldsymbol{\alpha}$ and fixed blendshape weight $\hat{\boldsymbol{\alpha}}_i$ corresponding to the $i$th key expression:
\begin{equation}
\begin{aligned}
\mathop{M_i}=\sum_{j=1}^{51}\boldsymbol{\alpha}^j\hat{\boldsymbol{\alpha}}_i^j{A_j}
\end{aligned}
\end{equation}
where $\boldsymbol{\alpha}^j$ is the $j$th element of $\boldsymbol{\alpha}$. $M_0$ is given by $M_0 = \max(0, 1-{\sum}_{i=1}^{19}M_i)$.

There are many existing performance driven facial animation methods generating blendshape weights with depth camera\cite{weise2011realtime, li2013realtime, bouaziz2013online} or single RGB camera\cite{cao20133d, cao2014displaced,chaudhuri2019joint}. As blendshape weights have semantic meaning, it's easy for artists to manually adjust the rigging parameters. %\red{(Typically, Our model can't directly use the expression parameter generated by other methods. We can use the parameter from Faceshift because our blendshapes are identical to them. Our basemodel fitting methods can also generate parameters for animation.)}
\section{Experiments}

\begin{figure*}[t]
\begin{center}
    \includegraphics[width=1.0\linewidth]{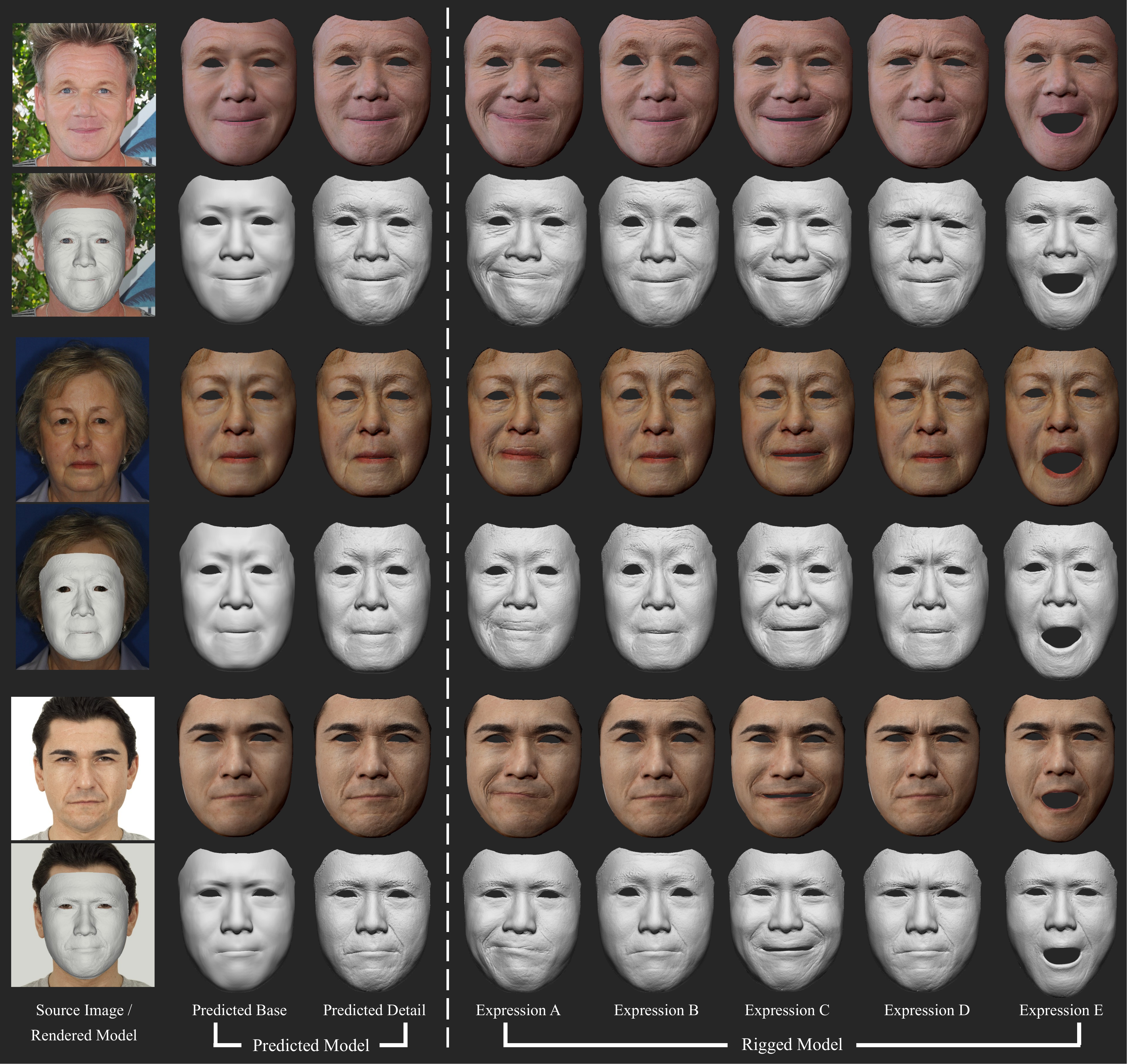}
\end{center}
    \vspace{-0.2in}
    \caption{We show our predicted faces in source expression and rigged expressions.  It is worth noting that the wrinkles in rigged expressions are predicted from the source image.}
\label{fig:results}
\end{figure*}

\begin{figure}[t]
\begin{center}
    \includegraphics[width=1.0\linewidth]{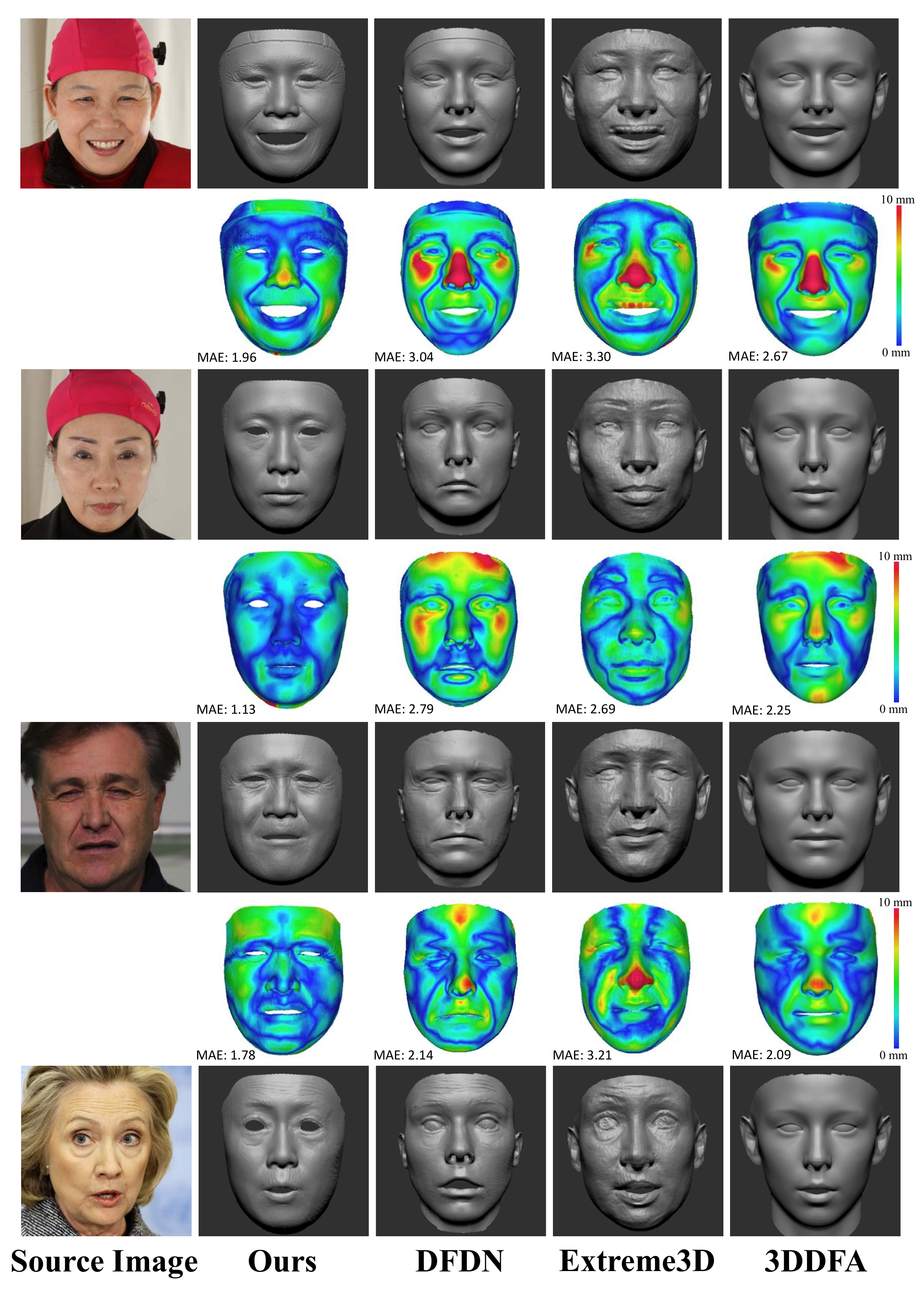}
\end{center}
    \vspace{-0.3in}
    \caption{Comparison of static 3D face prediction with previous methods.  The images in top two rows are from FaceScape, the image in third row is from volker sequence\cite{valgaerts2012lightweight}, and the image in bottom row is from Internet.  The top three images are with ground truth shapes, so we evaluate the reconstruction error and show the heat map below each row.  Our method predicts the lowest error comparing to three previous methods.}
\label{fig:compare}
\end{figure}

\subsection{Implement Detail}
%\red{Amount of training data, optimizer parameters, hyperparameters.}
We use 888 people in our dataset as training data with a total of 17760 displacement maps, leaving 50 people for testing. We use the Adam optimizer to train the network with learning rate as $2e^{-4}$. The input textures and output displacement maps' resolution of our network is both $1024 \times 1024$. We use 50 identity parameters, 52 expression parameters and 100 albedo parameters for our parametric model in all experiments.

\subsection{Evaluation of 3D Model Prediction}
The predicted riggable 3D faces are shown in Figure~\ref{fig:results}.  To show riggable feature of the recovered facial model, we rig the model to 5 specific expressions.
%, which is produced by FaceShift, a software which generate blendshapes parameters from captured depth maps.  
We can see the results of rigged models contain the photo-realistic detailed wrinkles, which cannot be recovered by previous methods.  The point-to-plane reconstruction error is computed between our model and the ground-truth shape. The mean error is reported in Table~\ref{tab:recon_error}.  
%We also show some failure cases of our method in Figure~\ref{fig:failure}.  
More results and the generated animations are shown in the supplementary material.

\begin{table}[]
\center 
\caption{3D face Prediction Error}
\begin{tabular}{lll}
\hline
method                    & mean error & variance \\ \hline
our method (all exp.)     & 1.39      & 2.33    \\ \hline
our method (source exp.)  & \textbf{1.22}      & \textbf{1.17}          \\
DFDN\cite{chen2019photo} (source exp.) & 2.19           & 3.20         \\
Extreme3D\cite{tran2018extreme} (source exp.) & 2.06           & 2.55         \\ 
3DDFA\cite{zhu2017face} (source exp.) & 2.17           & 3.23         \\ \hline
\end{tabular}
\label{tab:recon_error}
\end{table}

\subsection{Ablation Study}
\textbf{W/O dynamic detail.}  We try to use only one displacement map from source image for rigged expressions, and the other parts remain the same.  As shown in Figure~\ref{fig:ablation}, we find that the rigged model with dynamic detail shows the wrinkles caused by various expressions, which are not found in W/O dynamic method.

\textbf{W/O deforming Map.}  We change the input of our displacement map prediction network by replacing the deforming map with one-hot encoding for each of 20 target expressions.  As shown in Figure~\ref{fig:ablation}, we find the results without deforming map (W/O Def. Map) contain few details caused by expressions.

\subsection{Comparisons to Prior Works}
\begin{figure}[t]
\begin{center}
    \includegraphics[width=1.0\linewidth]{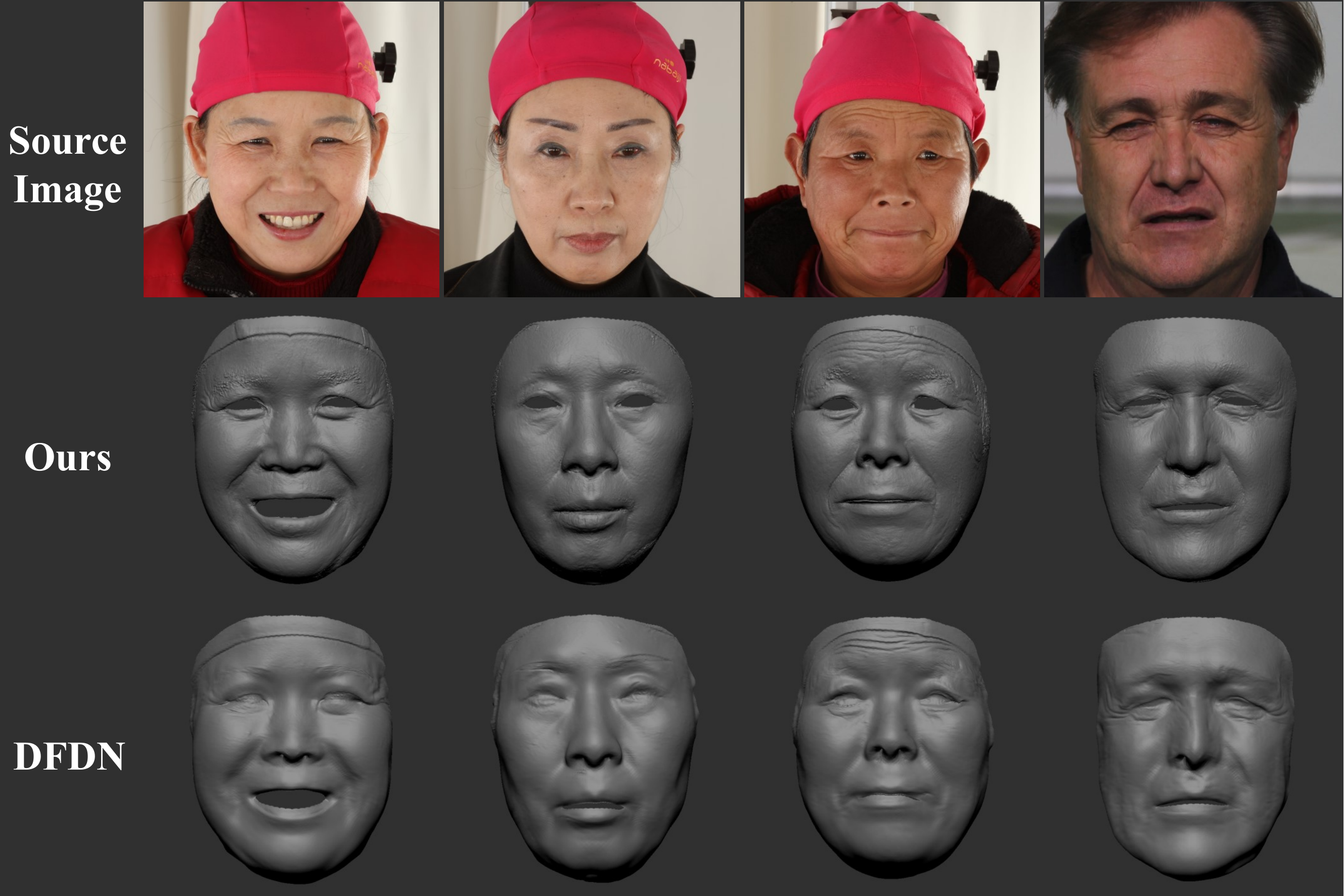}
\end{center}
    \vspace{-0.2in}
    \caption{Comparison of detail prediction. We adopt NICP\cite{amberg2007optimal} to register the base meshes of different methods to ground truth scans, and visualize the predicted details on common base meshes.}
\label{fig:orthogonal}
\end{figure}

\begin{figure}[t]
\begin{center}
    \includegraphics[width=1.0\linewidth]{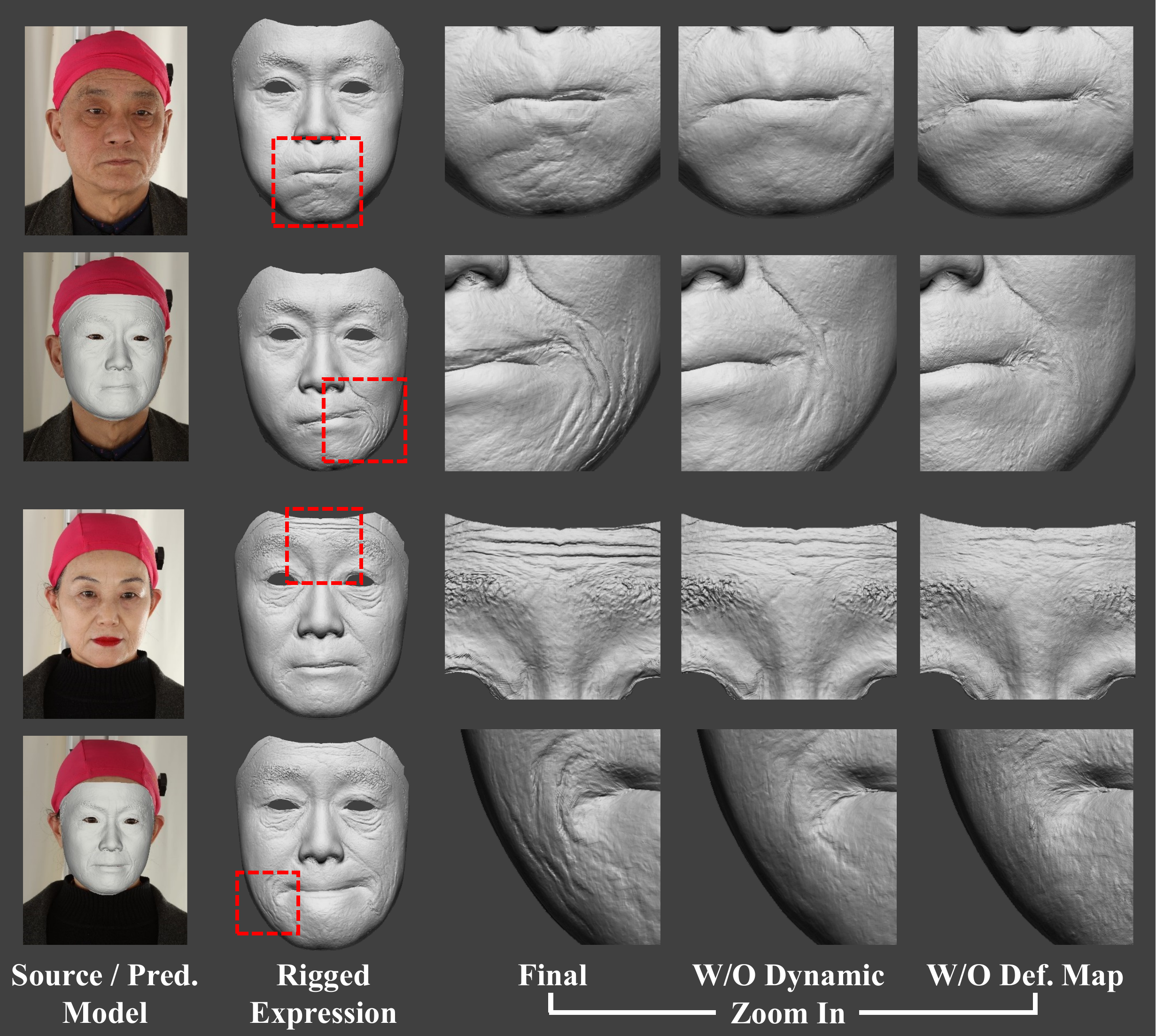}
\end{center}
    \vspace{-0.2in}
    \caption{Ablation study.  Our final model are able to recover wrinkles in rigged expressions, while the method W/O demforming map and W/O dynamic details cannot.}
\label{fig:ablation}
\end{figure}

We show the predicted results of our result and other works in Figure~\ref{fig:compare}. The comparison of detail prediction is shown in Figure~\ref{fig:orthogonal}. As most of the detailed face predicted by other works cannot be directly rigged to other expressions, we only show the face shape in the source expression.  Our results are visually better than previous methods, and also quantitatively better in the heat map of error.  We consider the major reason for our method to perform the best in accuracy is the strong representation power of our bilinear model, and the predicted details contribute to the visually plausible detailed geometry.

% \begin{figure}[t]
% \begin{center}
%     \includegraphics[width=1.0\linewidth]{figure/fig_heatmap.pdf}
% \end{center}
%     \vspace{-0.4in}
%     \caption{Heat map to visualize the prediction error.}
% \label{fig:heat_map}
% \end{figure}

\section{Conclusion}

We present a large-scale detailed 3D facial dataset, FaceScape. Comparing to previous public large-scale 3D face datasets, FaceScape provides the highest geometry quality and the largest model amount.
We explore to predict a detailed riggable 3D face model from a single image, and achieve high fidelity in dynamic detail synthesis.  We believe the release of FaceScape will spur the future researches including 3D facial modeling and parsing.
%A parametric model consists of registered base shapes and attached displacement maps and textures are generated.

\vspace{-1pt}
\section*{Acknowledgements}
\vspace{-1pt}
This work was supported by the grants -- NSFC 61627804 / U1936202, USDA 2018-67021-27416, JSNSF BK20192003, and a grant from Baidu Research.

\newpage

{\small
\bibliographystyle{ieee_fullname}
\bibliography{bib_face}
}

\newpage

\appendix
\section*{Supplementary}
\section{Animation}
We recommend watching the supplementary video, where the FaceScape dataset is briefly introduced and the generated animations are shown.  In the animation part, the 3D face model is predicted from a single wild image, then is rigged to the expressions captured by FaceShift\cite{weise2011realtime}.  As shown in the video, the face model predicted by our method can be rigged to various expressions while recovers the dynamic details, such as the wrinkles caused by expressions.  We also use the same rigging parameters to drive 3 different predicted models, and find that they appear different dynamic details.  This is because these details are related to the source subjects, not the rigging parameters.

\section{Model Processing Details}
The generation of topologically uniformed model has been briefly introduced in Section 3.2 of the main paper.  Here we supplement a detailed description of model registration and displacement map generation.

\textbf{Registration of base shape.}  We down-sample the raw recovered mesh into rough mesh with fewer triangle faces, namely base shape, and then build 3DMM for these simplified meshes. Firstly, the 2D landmarks are extracted from the frontal image, then the corresponding 3D landmarks are obtained by inverse-projection 2D landmarks.  The Procrustes transformation\cite{gower1975generalized} is used to register all landmarks to a standard 3D facial template with landmark annotations.  In this way, the pose and scale for all the scanned meshes are roughly aligned to the standard facial template.  Then we use Non-rigid ICP\cite{amberg2007optimal} to register the standard template mesh to scanned mesh in neutral expression. For scanned meshes in other 19 expressions, similar to \cite{cao2013facewarehouse}, the deformation transfer algorithm\cite{sumner2004deformation} is firstly used to deform the registered mesh in neutral expression to other expressions mimicking the deformation of a set of template meshes in corresponding expressions. Then the Non-rigid ICP\cite{amberg2007optimal} is used to register these deformed individual-specific templates to scanned meshes to fit the scans in non-neutral expressions more accurately.

\textbf{Displacement map generation.}  After obtaining the topology-uniformed base shape, we use displacement maps in UV space to represent middle and fine scale details that are not captured by the base model due to the small number of vertices and faces.  The most straightforward way to compute the displacement map is to calculate the distance from the surface of the registered model to the raw mesh. However, we find that there will be artifacts in the displacement map caused by the defects in the registration procedure. Thus the raw scan is firstly smoothed with Laplacian mesh smoothing. Then we trace the surface points of base mesh corresponding to pixels in the displacement map, and inverse-project the points to the raw mesh along normal direction to find its corresponding points.  The pixel value of the displacement map is set to the signed distance from the point on raw mesh to its corresponding point on the smoothed mesh.

\section{Base Model Fitting}

The base model fitting method has been briefly introduced in Section 4.1 of the main paper. Here we provide a detailed description of three parts in the objective function.

\textbf{Landmark Alignment.} Firstly the 2D landmarks $L$ are extracted from the image using an off-the-shelf facial landmark detector. Assuming the camera is weak perspective, the landmark alignment term is defined as the distance between the detected 2D landmark $L^{(k)}$ and its corresponding vertex projected on the image space:
\begin{equation}
\begin{aligned}
E_{lan}=||(sR(C_r \times \textbf{w}_{exp} \times \textbf{w}_{id})^{(k)}+\textbf{t})-L^{(k)}||^2_2
\end{aligned}
\end{equation}
where $s$ is the scale factor of the weak perspective function, $R$ is the rotation matrix and $\textbf{t}$ is the translation.

\textbf{Pixel Level Consistency.}
The pixel-level reconstruction term is used to match the geometry more accurately in the regions where no feature points such as cheeks exists. Under the assumption of Lambertian surfaces, we use the first three bands of Spherical Harmonics(SH)\cite{ramamoorthi2001signal} for illumination representation. The per-vertex albedo is represented as a PCA model based on our dataset with albedo parameter $\textbf{w}_{alb}$. The objective function is formulated as:
\begin{equation}
\begin{aligned}
E_{pixel}=\frac{1}{|\mathcal{V}|} \sum_{q \in \mathcal{V}}||\hat{I}(q)-I(q)||_{2}
\end{aligned}
\end{equation}
where $\mathcal{V}$ is the set of pixels corresponding to frontal vertices of the fitted mesh, $\hat{I}$ is the synthetic face, $I$ is the input image.

\textbf{Regularization.} 
We formulate the prior of identity, expression and albedo parameters as multivariate Gaussians around the average of our dataset for regularization. 
%Similarly to \cite{weise2011realtime}, the prior of expression parameter is formulated as Gaussian mixture model learned from a set of pre-generated expression parameters using \cite{weise2011realtime}.
The final objective function is given by:
\begin{equation}
\begin{aligned}
E=E_{lan}+\lambda_{1} E_{pixel}+{\lambda_2 E_{id}}+{\lambda_3 E_{exp}}+{\lambda_4 E_{alb}}
\end{aligned}
\end{equation}
where $E_{id}$, $E_{exp}$ and $E_{alb}$ are the regularization terms of expression, identity and albedo, respectively. $\lambda_1$, $\lambda_2$, $\lambda_3$ and $\lambda_4$ are the weights of different terms. We optimize the parameters alternatively. Following \cite{zhu2015high}, the vertex indices corresponding to contour landmarks of the face are updated after each iteration.

\section{Facial Capture System}
\begin{figure}[t]
\begin{center}
    \includegraphics[width=1.0\linewidth]{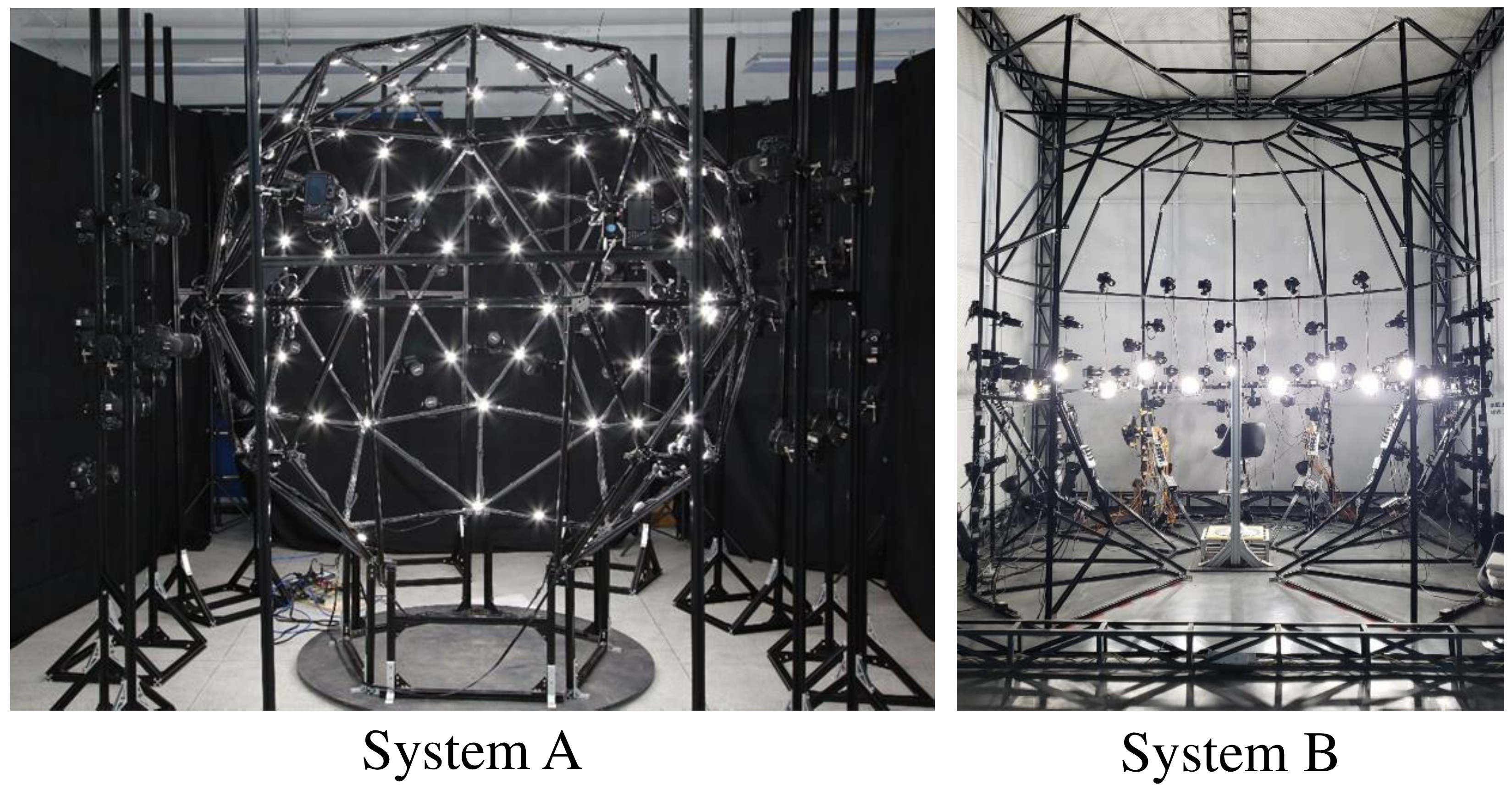}
\end{center}
    \caption{Our multi-view system to reconstruct the high quality detailed 3D face.  We captured the data in two different places, so there are two frameworks shown as system A and system B. }
\label{fig:hardware}
\end{figure}

The capturing system has been briefly introduced in Section 3.1 of the main paper.  Here we supplement the pictures of our system in Figure~\ref{fig:hardware}.  The system consists of the 68 DSLR camera array, controlled lighting and a centralized control sever.

\begin{figure}[t]
\begin{center}
    \includegraphics[width=1.0\linewidth]{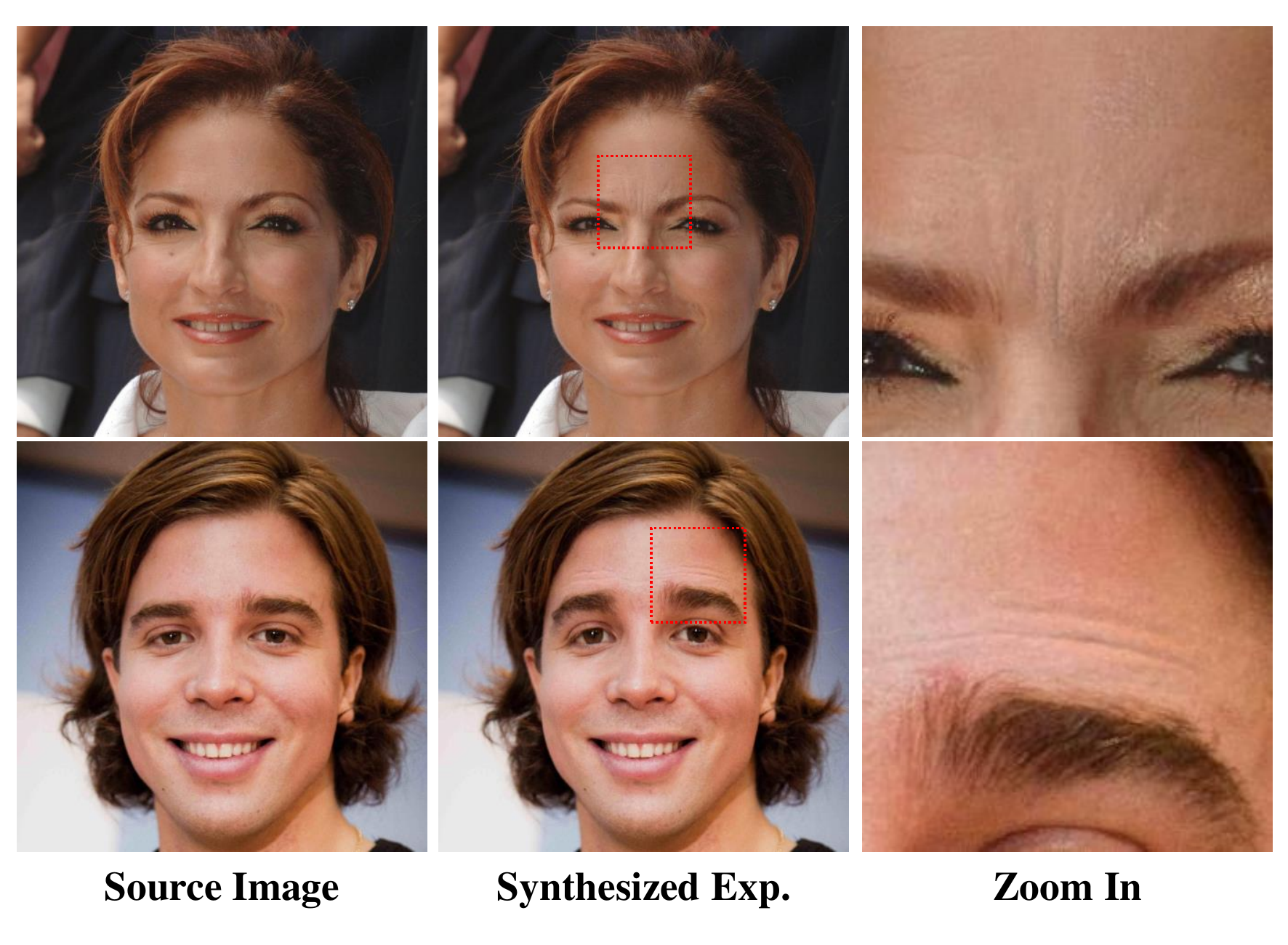}
\end{center}
    \caption{We use our recovered model for synthesizing images in another expression with detailed shading.}
\label{fig:synthesis}
\end{figure}

\begin{figure}[t]
\begin{center}
    \includegraphics[width=1.0\linewidth]{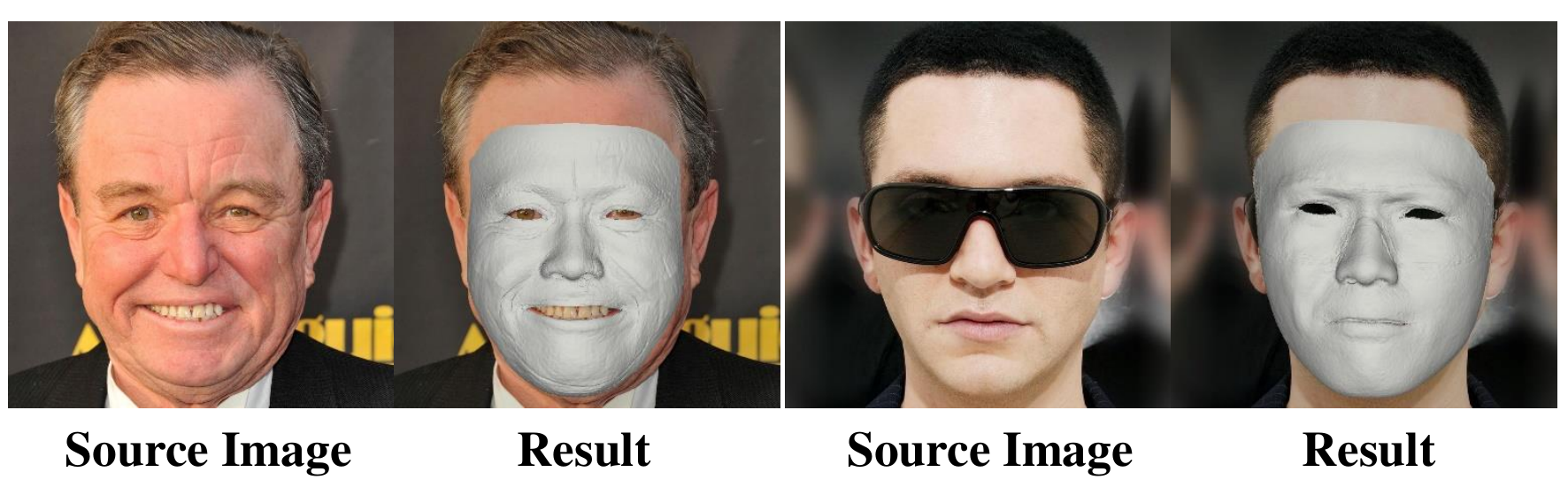}
\end{center}
    \vspace{-0.22in}
    \caption{Failure cases.  In the left, our prediction cannot recover the aquiline nose well, as this feature is not common in our dataset.  In the right, the wrong displacement map is predicted due to occlusion.}
\label{fig:failure}
\end{figure}

\section{More Results}

More results are supplemented in Figure~\ref{fig:more_results} as the extension of Figure 6 in our main paper.  It shows that our results recover 3D faces with photo-realistic details. The faces can be further rigged to other expressions, and the details in the new expressions are synthesized to make the rigged model plausible.

\begin{figure}[t]
\begin{center}
    \includegraphics[width=0.8\linewidth]{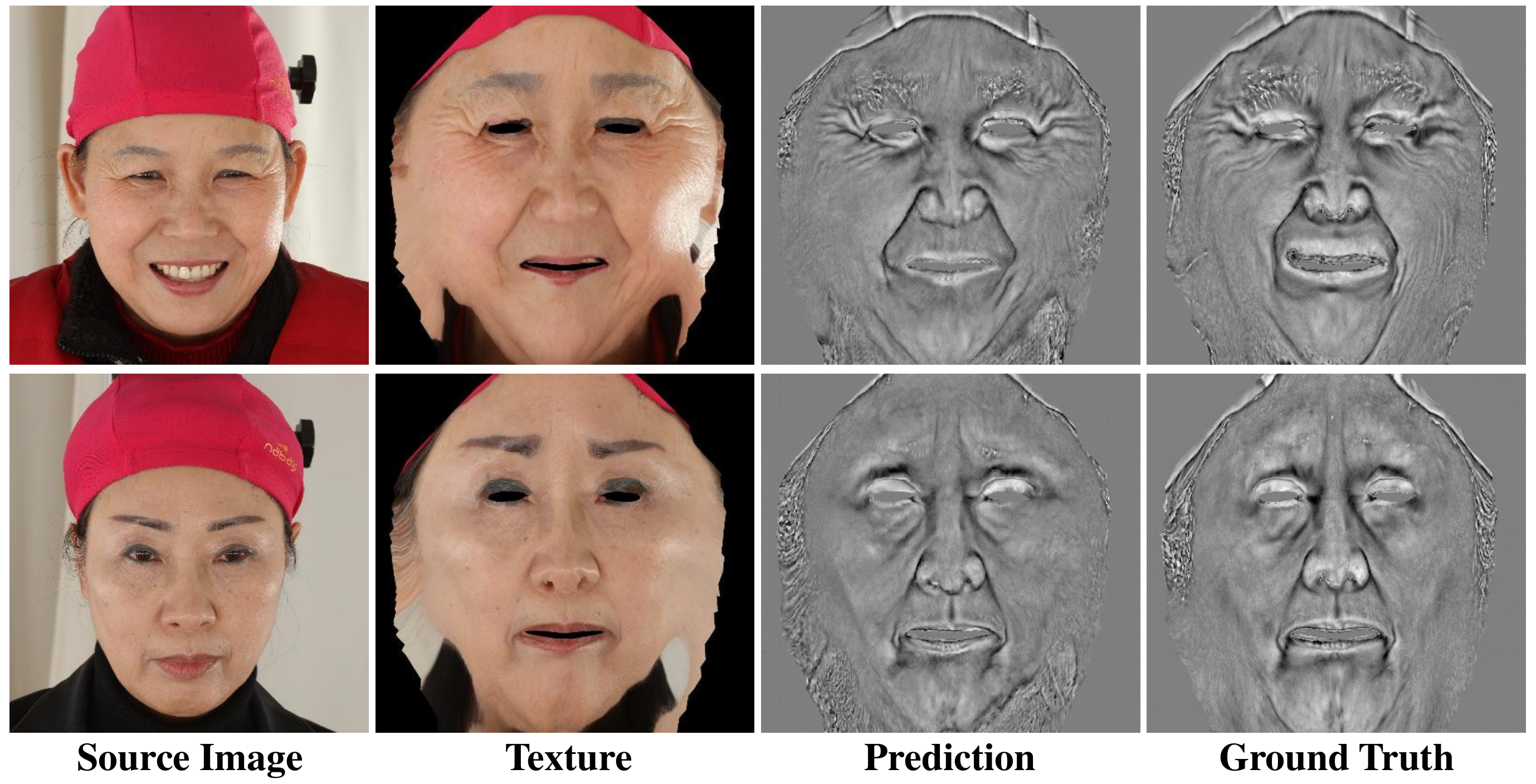}
\end{center}
    \vspace{-0.15in}
    \caption{Predicted displacement maps using our method and ground truth.}
    \vspace{-0.15in}
\label{fig:compare_dp}
\end{figure}

\begin{figure*}[t]
\begin{center}
    \includegraphics[width=1.0\linewidth]{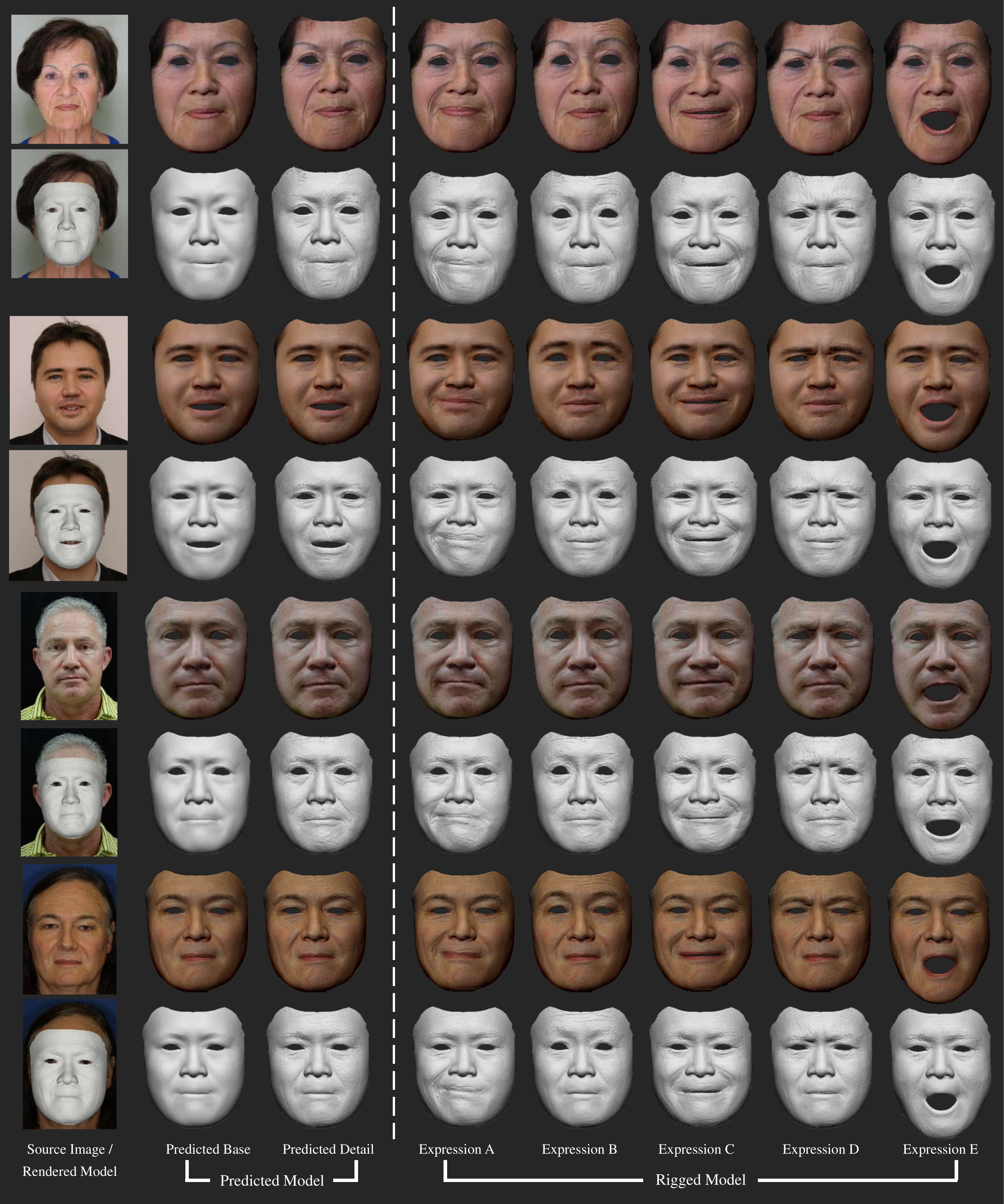}
\end{center}
    \caption{We show more results as the extension of Figure 6 in our main paper. }
\label{fig:more_results}
\end{figure*}

We supplement the comparison of the predicted and ground-truth displacement maps in Figure~\ref{fig:compare_dp} as the extension of Figure 7 in our main paper.

\section{More Models}
\begin{figure}[t]
\begin{center}
    \includegraphics[width=1.0\linewidth]{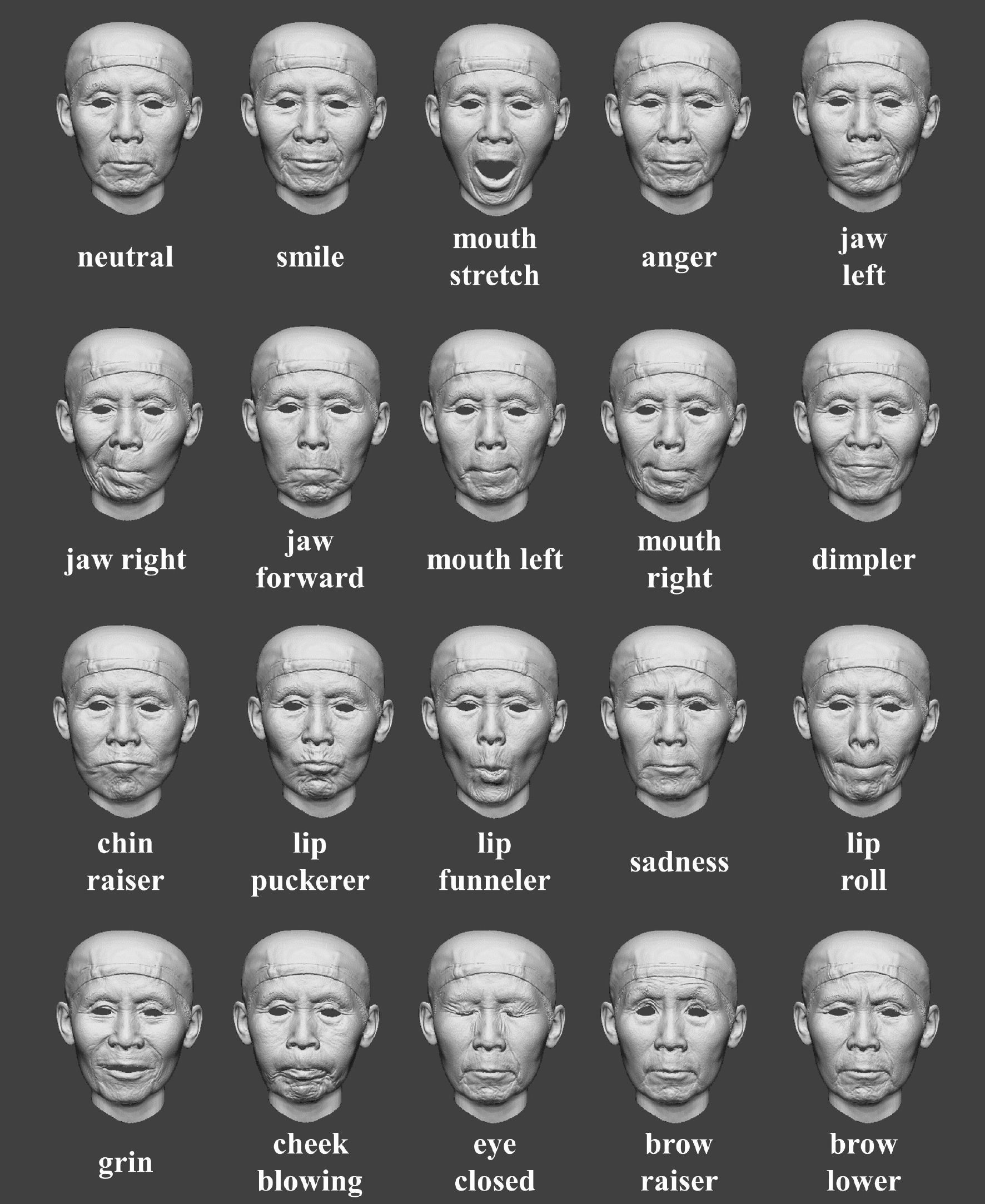}
\end{center}
    \caption{The 20 specified expressions which the subjects are asked to perform. }
\label{fig:more_exps}
\end{figure}

\begin{figure}[t]
\begin{center}
    \includegraphics[width=1.0\linewidth]{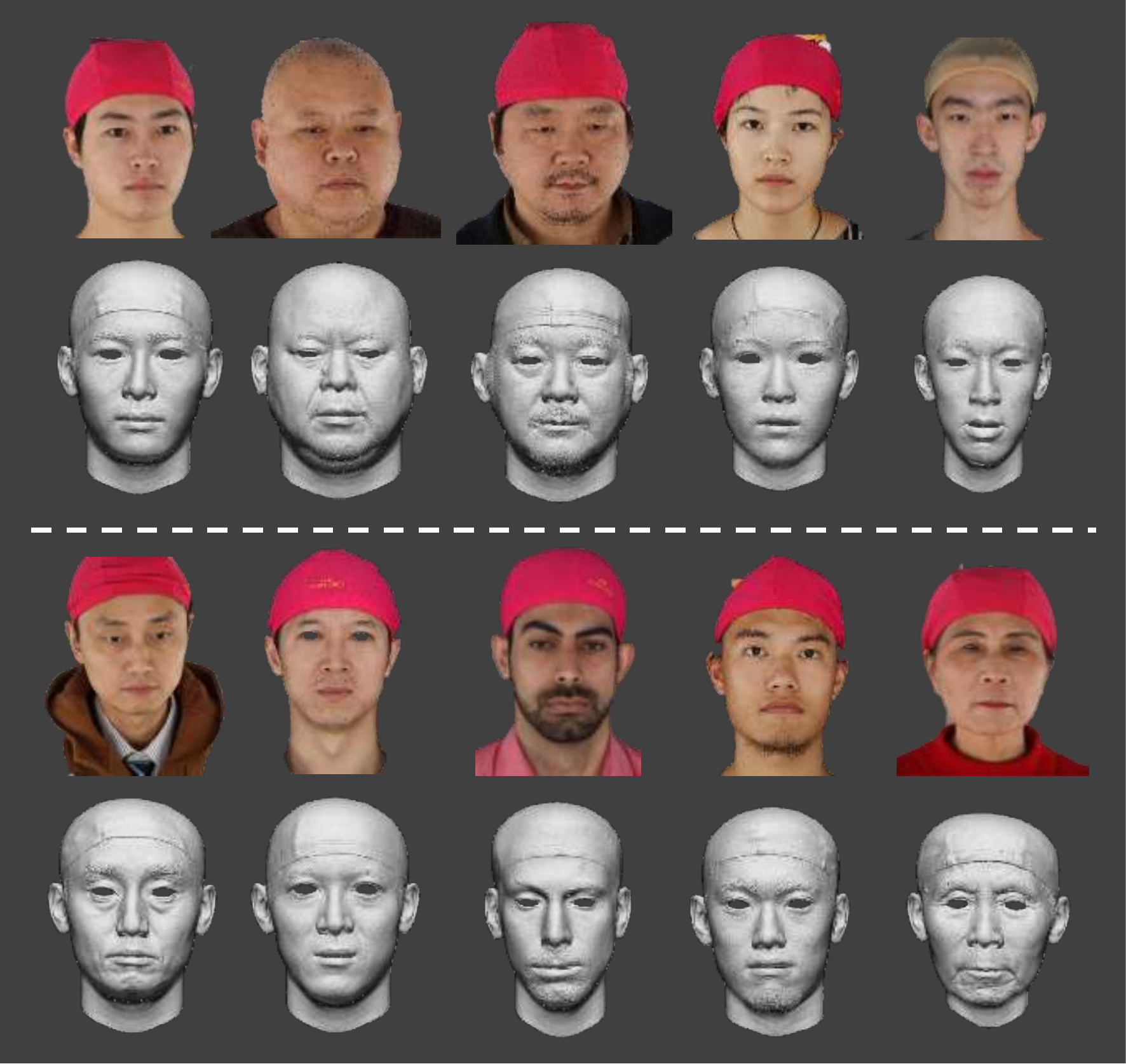}
\end{center}
    \caption{More models with different identities are shown in this figure.  The upper part is the images of the subjects, and lower part is the processed topologically uniformed models. }
\label{fig:more_id}
\end{figure}

We show the 20 captured expressions for each subject in Figure~\ref{fig:more_exps}, and show more subjects in neutral expression in Figure~\ref{fig:more_id}.  The diversity of models in expression and identity dimensions ensures the quality of bilinear face model generated on FaceScape dataset.

\section{Photo-realistic Image Synthesis} % application
%Our result model can be further used in photo-realistic facial manipulation. 
Similar to \cite{vlasic2005face,cao2013facewarehouse}, given a facial image, our bilinear model can be used to synthesize images in other expressions. Specifically, we use the base model fitting method to estimate the face model. Then we change the expression parameter to generate the face model in the target expression and warp the image pixels guided by translations of vertices on the 3D face model.  The details caused by the expression changing are further synthesized by adjusting the pixel shading. New pixel value is calculated based on the new normal from the predicted displacement map and estimated illumination in model fitting procedure. The synthesized images are shown in Figure~\ref{fig:synthesis}.

\section{Failure Case}
We show some failure cases of our method in Figure~\ref{fig:failure}.

\end{document}